\documentclass[12pt]{article}

\usepackage{times}
\usepackage{latexsym}

\usepackage{placeins}
\usepackage{array}
\usepackage{hyperref}
\usepackage{cleveref}

\usepackage[T1]{fontenc}

\usepackage[utf8]{inputenc}

\usepackage{microtype}

\usepackage{inconsolata}

\usepackage{graphicx}
\usepackage{spverbatim}
\usepackage{listings} 
\usepackage{fancyvrb}
\usepackage{fvextra} 
\usepackage{multirow}
\usepackage{booktabs}

\usepackage{arxiv}

\usepackage[utf8]{inputenc} 
\usepackage[T1]{fontenc}    
\usepackage{hyperref}       
\usepackage{url}            
\usepackage{booktabs}       
\usepackage{amsfonts}       
\usepackage{nicefrac}       
\usepackage{microtype}      
\usepackage{lipsum}
\usepackage{graphicx}

\usepackage[authoryear]{natbib}
\graphicspath{ {./images/} }

\title{To Words and Beyond: Probing Large Language Models for Sentence-Level Psycholinguistic Norms of Memorability and Reading Times}

\author{
 \textbf{Thomas Hikaru Clark\textsuperscript{1}},
 \textbf{Carlos Arriaga\textsuperscript{2}},
 \textbf{Javier Conde\textsuperscript{2}},
\\
 \textbf{Gonzalo Mart{\'i}nez\textsuperscript{2}},
 \textbf{Pedro Reviriego\textsuperscript{2}}
\\
 \textsuperscript{1}Massachusetts Institute of Technology
\\
 \textsuperscript{2}Universidad Polit{\'e}cnica de Madrid,
\\
 \small{
   \textbf{Correspondence:} \href{mailto:thclark@mit.edu}{thclark@mit.edu}
 }
}

\begin{document}
\maketitle

\begin{abstract}
Large Language Models (LLMs) have recently been shown to produce estimates of psycholinguistic norms, such as valence, arousal, or concreteness, for words and multiword expressions, that correlate with human judgments. These estimates are obtained by prompting an LLM, in zero-shot fashion, with a question similar to those used in human studies. Meanwhile, for other norms such as lexical decision time or age of acquisition, LLMs require supervised fine-tuning to obtain results that align with ground-truth values. In this paper, we extend this approach to the previously unstudied features of sentence memorability and reading times, which involve the relationship between multiple words in a sentence-level context. Our results show that via fine-tuning, models can provide estimates that correlate with human-derived norms and exceed the predictive power of interpretable baseline predictors, demonstrating that LLMs contain useful information about sentence-level features. 
At the same time, our results show very mixed zero-shot and few-shot performance, providing further evidence that care is needed when using LLM-prompting as a proxy for human cognitive measures. 

\end{abstract}
\FloatBarrier

\section{Introduction}

How much useful information do Large Language Models (LLMs) contain about human psycholinguistic features? Prior work indicates that LLMs are able to predict word-level features such as concreteness or valence \cite{trottCanLargeLanguage2024}, age of acquisition \cite{sendin2025combiningageacquisition}, and lexical decision times \cite{martinezSimulatingLexicalDecision2025} as well as for multi-word expressions \cite{martinez2024usingmultiword}, \cite{brysbaert2024movingfamiliarity}. 
One method of testing for the presence of useful psycholinguistically relevant information within LLMs is to simply prompt LLMs by asking a psycholinguistic query directly in zero-shot fashion. However, some evidence points to a lack of \textbf{introspection} in LLMs --- their responses to prompts are not necessarily consistent with their latent knowledge that can be accessed in other ways, such as by inspecting token log-probabilities rather than directly prompting \cite{songLanguageModelsFail2025, huPromptingNotSubstitute2023}. 
Fine-tuning language models based on small amounts of supervised data provides a way to better capitalize upon the rich learned model representations of pre-trained models. For example, \cite{conde2025adding} fine-tuned Llama 3 models to predict English familiarity ratings, achieving Pearson's correlation improvements of up to 0.3 over zero-shot baselines. 

\begin{figure*}[ht]
  \centering
  \includegraphics[width=0.95\textwidth]{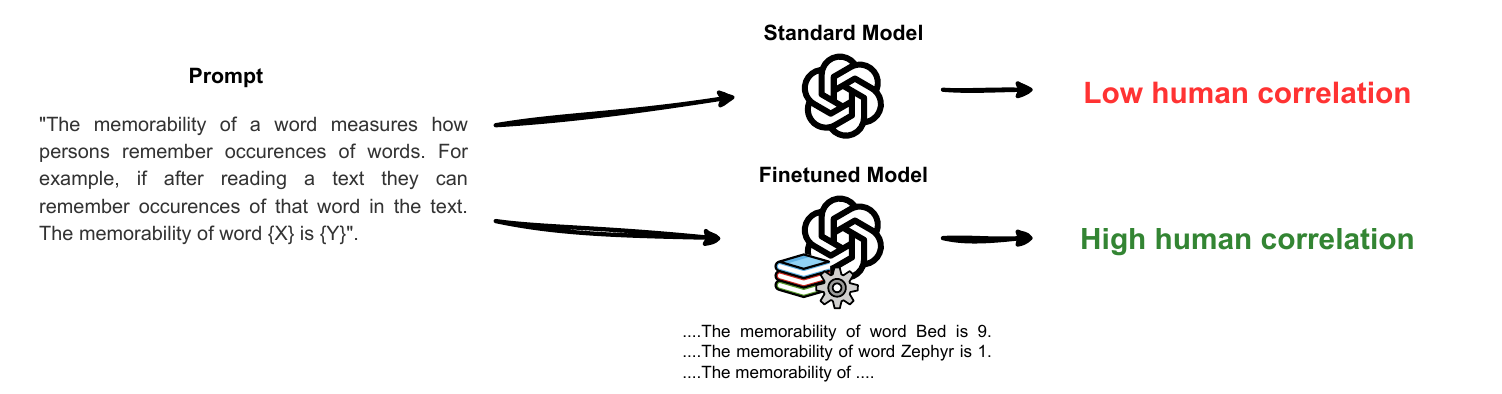}
  \caption{Overview of our approach, contrasting zero-shot prompting with fine-tuning on small supervised datasets for predicting psycholinguistic norms.}
  \label{fig:overview}
\end{figure*}

The approach of prompting an LLM to provide psycholinguistic norms differs from approaches which directly predict psycholinguistic features based on theoretically motivated, interpretable features. For example, surprisal theory states that human processing difficulty during reading is related to the predictability of words in context \cite{haleProbabilisticEarleyParser2001, smithEffectWordPredictability2013, levyExpectationbasedSyntacticComprehension2008}. For the domain of word memorability, \cite{tuckuteIntrinsicallyMemorableWords2025} used the theoretically motivated predictors of number of meanings and number of synonyms to predict the memorability of words, while \cite{clarkDistinctiveMeaningMakes2026} used the distinctiveness of a sentence's Sentence-BERT semantic embedding \cite{reimersSentenceBERTSentenceEmbeddings2019}, as a theoretically motivated, zero-shot, sentence-level predictor of sentence memorability (among other features such as average word memorability and average word frequency). In our view, these methods address different questions, and thereby form a complementary set of approaches that collectively shed light on the functioning of LLMs and their alignment with human cognition. 

Probing for psycholinguistic features exists as part of a broader paradigm of using artificial intelligence models to study and simulate human cognition \cite{frankCognitiveModelingUsing2025, hagendorffMachinePsychology2024,binzFoundationModelPredict2025}. 
The use of LLM-prompting in cognitive science has also raised questions both about the best methodologies for eliciting predictions, as well as calls for caution regarding using LLM outputs as a substitute for human data \cite{gaoTakeCautionUsing2025, dentellaSystematicTestingThree2023}. Within this landscape, it is crucial to understand how well LLM outputs actually align with varied psycholinguistic features, whether and how this alignment can be improved, and whether the paradigm of LLM-prompting can be extended to sentence-level, and not just word-level, features. 

In this paper, we go beyond existing work studying word-level psycholinguistic features in LLMs, and explore whether LLMs can generate predictions for norms which require processing at the sentence level. Specifically, we systematically compare zero-shot vs. fine-tuned predictions from LLMs to psycholinguistic behavioral norms, across the domains of memorability and reading times.   

Our contributions are as follows: 

\begin{enumerate}
    \item We demonstrate that for memorability (both word and sentence-level), zero-shot prompting yields predictions that are uncorrelated with empirical norms. Meanwhile, for reading times, zero-shot prompting yields predictions that do correlate with empirical norms --- modestly for self-paced reading and more strongly for eye-tracking data.
    \item We demonstrate that, for both memorability and reading times, fine-tuning on a few hundred examples yields predictions that are strongly correlated with empirical norms --- exceeding the predictive power of simple, interpretable baselines. This shows that via fine-tuning, the prompting of LLMs for psycholinguistic norms can be extended to sentence-level norms, not just word-level features.
\end{enumerate}

The remainder of the paper is organized as follows. Section 2 discusses the data, models, and procedures used in the evaluation. Section 3 presents the results, which are then discussed in Section 4. We conclude by discussing implications for the intersection of psycholinguistics and NLP, and the limitations and caveats of this approach. 


\section{Methods}

This section describes the data, models, and procedures used to generate and evaluate LLM predictions of memorability and reading times, across zero-shot and fine-tuned settings. A general overview of the procedure is illustrated in Figure \ref{fig:overview}.

\subsection{Data}

Here we introduce the datasets which we use in our model evaluations. We focus on two different but important psycholinguistic domains --- memorability and reading times --- which have not yet been the target of comparison with LLM predictions. In contrast to past work  \cite{condePsycholinguisticWordFeatures2025, trottCanLargeLanguage2024}, which has considered word-level features such as the Glasgow word norms \cite{scottGlasgowNormsRatings2019a} or the Lancaster sensorimotor word norms \cite{lynottLancasterSensorimotorNorms2020},
the domains of sentence memorability and reading times pose a distinct challenge for estimating psycholinguistic norms via LLM prompting, because they are \textit{sentence-level} features. For instance, the memorability of a sentence is a property of the entire sentence, influenced by its compositional meaning \cite{clarkDistinctiveMeaningMakes2026}. Likewise, while reading times are known to depend on certain stable properties of a word, such as its length, frequency, or age of acquisition \cite{smithEffectWordPredictability2013, brothersWordPredictabilityEffects2021, lukeProvoCorpusLarge2018, dembergDataEyetrackingCorpora2008, kennedyFrequencyPredictabilityEffects2013}, they also depend on its contextual surprisal, which varies from sentence to sentence. Any model that successfully predicts reading time variation across different instances of the same wordform will need to take into account the relationship between multiple words in a sentence.

We also include word memorability as a comparison for sentence memorability. For both types of memorability data, norms are collected not by asking human raters for their judgments, but rather by conducting a repeat detection experiment; this makes these norms inherently more scarce and expensive to acquire. 
Another challenge of memorability data is its dissociation from human subjective judgments, as reported in prior literature. \cite{isolaWhatMakesPhotograph2014} report that subjective human judgments of the memorability of images actually correlate \textit{negatively} with empirically measured memorability, while \cite{clarkDistinctiveMeaningMakes2026} report that subjective memorability judgments for sentences have a weak correlation of 0.24 with empirical sentence memorability. Therefore, for all of the following datasets --- word memorability, sentence memorability, and reading times --- we have reason to believe that LLM predictions may not align as well with the ground truth compared to results from the prior literature. 

\subsubsection{Word Memorability}

We use the word memorability data of \cite{tuckuteIntrinsicallyMemorableWords2025}\footnote{Experiment 1, available at \url{https://github.com/gretatuckute/memorable_words} under MIT license.}, which contains 2109 English words with empirical memorability scores collected via a behavioral experiment with native English speakers. This experimental paradigm builds on a body of work related to measuring the intrinsic memorability of various classes of stimuli, such as images \cite{isolaUnderstandingIntrinsicMemorability2011} and faces \cite{bainbridgeMemorabilityPeopleIntrinsic2017}. A word's memorability is quantified as the average accuracy of responses across participants within a repeat detection (recognition memory) paradigm, yielding values between 0 and 1. In this paradigm, reporting a novel stimulus as familiar and failing to report a repeat stimulus as familiar are the two sources of incorrect responses. 

\subsubsection{Sentence Memorability}

We use the sentence memorability data of \cite{clarkDistinctiveMeaningMakes2026}\footnote{Available at \url{https://github.com/thomashikaru/sentence_memorability_share} under MIT license.}, which contains 2500 English sentences with empirical memorability scores collected via a behavioral experiment with native English speakers, following the same experimental paradigm as \cite{tuckuteIntrinsicallyMemorableWords2025}, thereby resulting in scores between 0 and 1. \ref{tab:sentmem-examples} shows examples of high- and low-memorability sentences from the dataset.  

\begin{table}[htb]
  \centering
  {\small
  \begin{tabular}{|w{l}{4.5cm}|w{c}{2cm}|}
    \hline
    \textbf{Sentence} & \textbf{Memorability} \\
    \hline
    Does olive oil work for tanning?  & 0.98 \\
    Scott cried, pursing his pink lips.  & 0.89 \\
    The weather was warm and dry.  & 0.79 \\
    I can't get hold of him.  & 0.72 \\
    We want to make it better.  & 0.56 \\
    \hline
  \end{tabular}
  }
  \caption{Example sentences from the dataset of \cite{clarkDistinctiveMeaningMakes2026}, with human-derived memorability scores. }
  \label{tab:sentmem-examples}
\end{table}

\subsubsection{Self-Paced Reading Times}

We use the reading time data from the Natural Stories Corpus \cite{futrellNaturalStoriesCorpus2021}\footnote{Available at \url{https://github.com/languageMIT/naturalstories} under CC BY-NC-SA 4.0 license.}, which used self-paced reading \cite{aaronsonPerformanceTheoriesSentence1976, mitchellEffectsContextContent1978} to gather average reading durations for words in naturalistic stories. In this paradigm, speakers are presented with one word at a time on a screen, and advance to the next word by pressing a key. Reading duration (in milliseconds) is the time between key presses (while the target word is on screen). The dataset consists of 433 sentences comprising 10,256 words. 

\subsubsection{Eye-Tracking Reading Times}

We use the reading time data from the OneStop Corpus, which used eye-tracking to gather reading times for words in a variety of news articles \cite{berzakOneStop360ParticipantEnglish2025}\footnote{Available at \url{https://osf.io/2prdq/} under CC BY 4.0 license.}. In this paradigm, speakers read naturalistically while their eye movements are tracked, yielding reading measures with high spatial and temporal resolution. Multiple reading measures are available, including first fixation duration, gaze duration, and total duration. For our main reading measure in this study, we use gaze duration (the sum, in milliseconds, of the durations of all fixations that land on a word during the first pass, before the gaze leaves the word), as this has an intuitive interpretation as the amount of time used to read a word when it is first encountered. In a different dataset \cite{siegelmanExpandingHorizonsCrosslinguistic2022}, gaze duration was shown to be largely predictable based on word length, frequency, and surprisal, with $R^2$ values of 0.6-0.8 \cite{opedalRoleContextReading2024}. The dataset consists of 1,213 sentences comprising 36,120 words. 

\subsection{Model Evaluation}

Here we describe in detail our procedure for extracting predictions from models in three settings: zero-shot, few-shot and fine-tuning.
In line with the majority of previous work that has relied on models from the GPT-4 family, we used the \verb|GPT-4o-mini-2024-07-28| model \cite{conde2025adding}. 
Additionally, we test models from the Llama, Gemma, and Qwen families to provide a comparison of model performance. The selected models are Llama-3.1-8b-Instruct, Gemma-3-27b-it, and Qwen3-32b. Due to hardware limitation during fine-tuning, Qwen3-32b and Gemma-3-27b-it were loaded in a 8 bit resolution rather that the full 16 bit resolution

\subsubsection{Zero-Shot}
For the zero-shot evaluation, we use OpenAI's batch API to submit all requests. 
For all requests in both zero-shot and fine-tuned model evaluation, the temperature is set to 0; this forces the model to use greedy sampling, where it always selects the token with the highest probability. This generates nearly deterministic \cite{he2025nondeterminism} outputs, which facilitates reproducibility.

We use the following zero-shot prompt for the word-memorability dataset: 

\begin{lstlisting}[]
You are an expert in psycholinguistics. Your task is to estimate the memorability of English words. You will give each word a rating from 0 to 1 with two decimal digits. A rating of 1 indicates that the word is maximally memorable, meaning that people who see the word always remember having seen it later, and never confuse it with a different word (even a similar one). A rating of 0 indicates that the word is not memorable, meaning that people who see it forget it or may confuse it with another word. Please limit your answer to a number with two decimal digits. The word is {word}
\end{lstlisting}

For each entry in the dataset, \verb|{word}| was replaced by the corresponding word of the dataset. 

For sentence memorability, a similar prompt was used, simply replacing references to ``words'' with references to ``sentences''. 

For both self-paced and eye-tracking reading times, the requested output was a JSON-like structure containing the estimates for each word of the sentence. 
\begin{lstlisting}[]
You are an expert in psycholinguistics. Your task is to estimate how long, in milliseconds, an average reader will take to read each word of an English sentence. Take into accounts factors such as the difficulty of reading the word and the context of the word within the sentence. Output a JSON-like data structure containing word-duration pairs. For example, for the sentence ''I like cats'' and the reading time estimates 100ms, 200ms, 200ms, the output must be {'I':200, 'like': 200, 'cats': 200}. Include duplicate keys if there are duplicate words even if the result is not strictly valid JSON. The order of keys should be identical to the order of words in the sentence. Do not add any other information. The sentence is: {sentence}
\end{lstlisting}

For sentence-level data, we align the model's output with the input while accounting for the possibility of missing or inserted words using a dynamic programming approach, implemented via the \verb|jiwer| Python library. 

\subsubsection{Few-Shot}
Few-shot evaluation proceeds identically to zero-shot evaluation, with the difference that three supervised examples are provided as part of the prompt. 

\subsubsection{Fine-Tuning}
We performed Supervised Fine-Tuning (SFT) in which we provided the model with both the prompt and the expected output based on the ground truth. For all fine-tuning processes, 25\% of the dataset was used for training and 75\% for evaluation. For the word and sentence memorability datasets, the expected output was the raw estimate (a single number), whereas for the reading time datasets, the expected output was a JSON-like structure (as described above). During the training process, the model weights are adjusted to adapt the estimate to the expected value. 

To train GPT-4o, we used the default fine-tuning API settings, with a fixed learning rate of 1.8. Once the training is completed, the fine-tuned version is stored and can be queried later through the batch API. The other models were fine-tuned locally using LoRA (Low-Rank Adaptation), a technique to adapt large models to specific tasks by training only a small number of new parameters (low-rank adapters) instead of the entire model. The details of the fine-tuning process are presented in Table \ref{tab:fineDetails}.

The prompts used for fine-tuning were the same ones used for the zero-shot evaluation. Once a fine-tuning job is created, the training process begins until the performance converges. For all evaluations, the final model checkpoint (after 3 epochs of fine-tuning) was used. 

\begin{table}[htb]
    \small
  \centering
  \begin{tabular}{|p{2.5cm}|c|c|c|}
    \hline
    \textbf{Dataset} & \textbf{\# Examples} & \textbf{Epochs} & \textbf{Batch Size} \\
    \hline 
    Word mem. & 527 & \multirow{4}{*}{3} & \multirow{4}{*}{1} \\
    \cline{1-2}
    Sentence mem. & 625 &  &  \\
    \cline{1-2}
    Self-paced & 108 &  &  \\
    \cline{1-2}
    Eye-tracking & 303 &  &  \\
    \hline
  \end{tabular}%
  \caption{Fine-tuning parameters used for each dataset. Note that for the reading time datasets, ``\# Examples'' denotes the number of sentences.}
  \label{tab:fineDetails}
\end{table}

Once the tuning process was completed, we evaluated its performance with the test dataset using the same methodology as in the zero-shot evaluation. 

\subsection{Correlation Analysis}

For each set of model predictions, we evaluate the correlation between human behavioral measures and the model-generated values. 
We quantify the correlation using the Pearson correlation coefficient and $R^2$ values. 

\subsection{Baselines}

In order to compare the performance of model predictions for the norms of memorability and reading times, we establish baselines using linear regressions trained on a) interpretable features from the literature, argued to strongly correlate with each psycholinguistic feature of interest, and b) word and sentence embeddings. Using 100 random 0.75/0.25 train/test splits of the data, we fit regressions using each baseline predictor on the training data, and then evaluate the $R^2$ value on the held-out test data. This procedure yields a distribution over $R^2$ values, from which we report the mean.

For word memorability, we use the number of meanings, number of synonyms, and word frequency, using values provided in the dataset of \cite{tuckuteIntrinsicallyMemorableWords2025}. In that study, the values for number of synonyms and number of meanings were collected via a human norming study, and word frequency was computed using the Subtlex corpus \cite{brysbaertMovingKuceraFrancis2009}. 
Additionally, we include a regression based on all three scalar predictors, as well as the 300-dimensional GloVe embedding for each word.
For sentence memorability, we use Sentence-BERT embedding distinctiveness, average word memorability, and average word frequency, using values provided in the dataset of \cite{clarkDistinctiveMeaningMakes2026}, since these were identified as predictors of sentence memorability. In that study, Sentence-BERT distinctiveness was computed as the mean cosine distance of a sentence's Sentence-BERT representation to all other sentences in a large and diverse sample of sentences, while average word memorability was estimated using the human norms of \cite{tuckuteIntrinsicallyMemorableWords2025} and word frequency was taken from the Subtlex corpus and averaged across words in a sentence.
Additionally, we include a regression based on all three scalar predictors, as well as the 384-dimensional Sentence-BERT embedding for each sentence. 

For the reading time norms, we establish baselines using the interpretable features of word length, frequency, and contextual surprisal, following past work \cite{dembergDataEyetrackingCorpora2008,brothersWordPredictabilityEffects2021,smithEffectWordPredictability2013, wilcoxTestingPredictionsSurprisal2023,opedalRoleContextReading2024}. Word length was defined simply as the number of characters in the orthographic representation of the word. Frequency was computed using the \texttt{wordfreq} package \cite{speerRspeerWordfreqV302022} for Python. Surprisal was computed using the \texttt{GPT-2} language model \cite{radfordLanguageModelsAre2019} and the \texttt{wordsprobability} Python package \cite{pimentelHowComputeProbability2024}.
Additionally, we include linear regressions based on all three scalar predictors, as well as 768-dimensional contextual BERT embeddings and 300-dimensional non-contextual GloVe embeddings. For words which decompose into multiple BERT tokens, only the first token is used. 

\section{Results}

This section presents and briefly discusses the results for each of the datasets. The Llama model did not succeed at responding consistently to the standard prompts, instead sometimes writing code or producing other non-usable output; Llama results are thus omitted from the following analysis, but we note that our approach therefore does not work for all LLMs. 

\subsection{Word Memorability}

\begin{figure}[htb]
\centering
  \includegraphics[width=0.8\columnwidth]{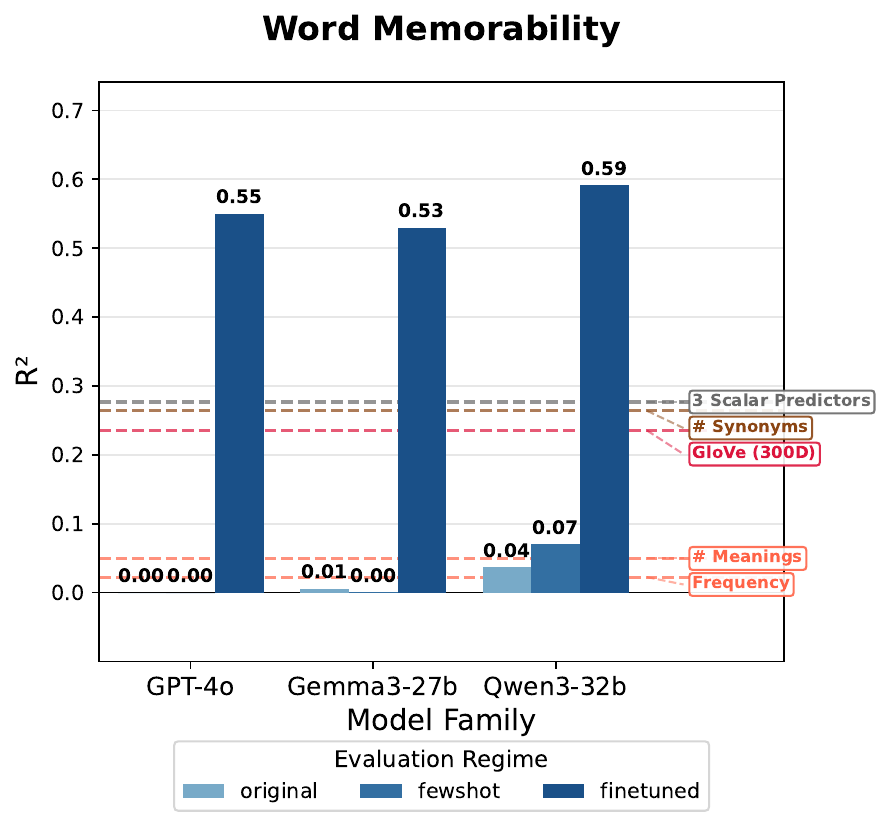}
  \caption {The correlation between model predictions and ground truth norms for word memorability, across 3 model families and in 3 evaluation regimes. For comparison, the mean correlation with the predictions of interpretable baseline predictors is included.}
  \label{fig:word-mem}
\end{figure}

Zero-shot and few-shot model predictions are essentially uncorrelated with empirical word memorability scores, while fine-tuned models achieve $R^2$ values of $0.53 \sim 0.59$ (\ref{fig:word-mem}). This exceeds the mean $R^2$ of 0.28 using the combined baseline predictors of number of meanings, number of synonyms, and frequency. 

\subsection{Sentence Memorability}
\begin{figure}[htb]
\centering
  \includegraphics[width=0.8\columnwidth]{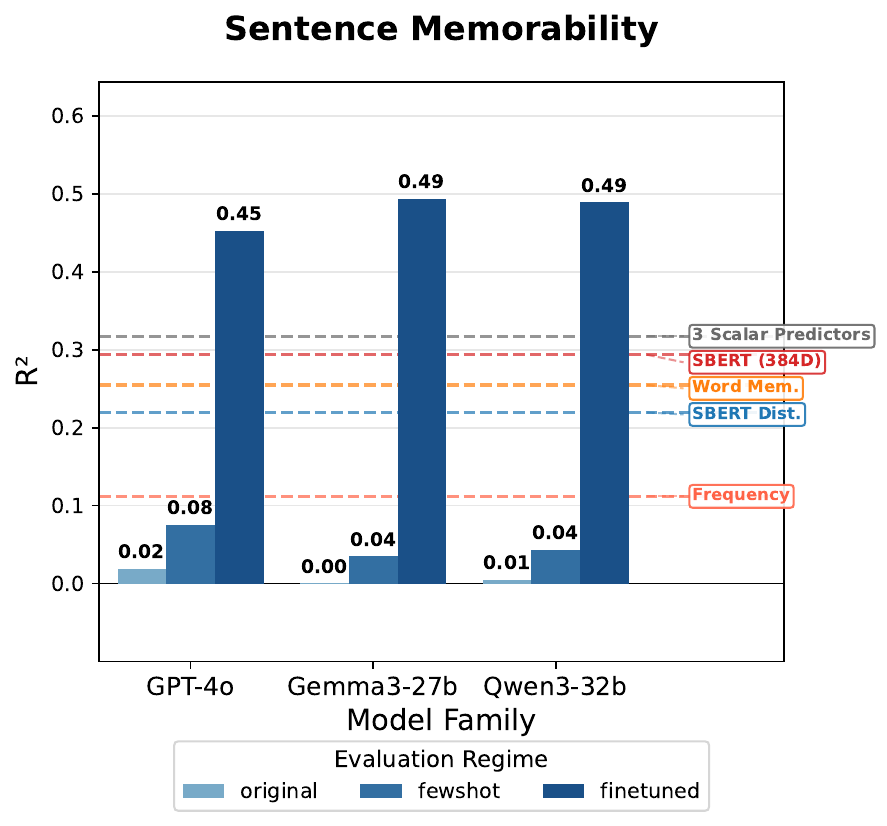}
  \caption {The correlation between model predictions and ground truth norms for sentence memorability, across 3 model families and in 3 evaluation regimes. For comparison, the mean correlation with the predictions of interpretable baseline predictors is included.}
  \label{fig:sent-mem}
\end{figure}

Zero-shot and few-shot model predictions are essentially uncorrelated with empirical sentence memorability scores, while fine-tuned models achieve $R^2$ values of $0.45 \sim 0.49$ (\ref{fig:sent-mem}). This exceeds the mean $R^2$ of 0.32 using the combined baseline predictors of Sentence-BERT distinctiveness, average word-level memorability, and average word frequency. 

\subsection{Reading Times}

\begin{figure}[htb]
\centering
  \includegraphics[width=0.8\columnwidth]{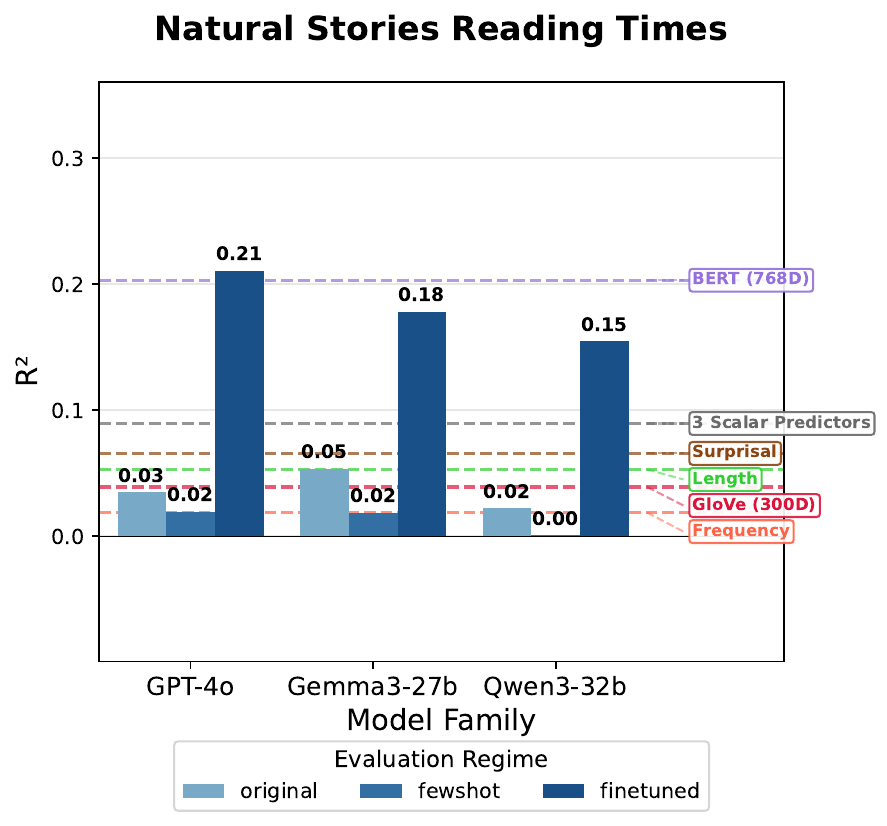}
  \caption {The correlation between model predictions and ground truth norms for self-paced reading times (Natural Stories corpus), across 3 model families and in 3 evaluation regimes. For comparison, the mean correlation with the predictions of interpretable baseline predictors is included.}
  \label{fig:rt-natural-stories}
\end{figure}

For the Natural Stories Corpus, zero-shot model predictions have low correlations of $0.02 \sim 0.05$ with empirical sentence memorability scores, while fine-tuned models achieve $R^2$ values of $0.15 \sim 0.21$ (\ref{fig:rt-natural-stories}). This is considerably lower than the $R^2$ values attained by fine-tuned models on the memorability data, but higher than the correlation of 0.08 using the combined scalar baseline predictors, and on par with the $R^2$ value achieved by predicting values using a linear model trained on words' 768-dimensional contextual BERT embeddings. 

\begin{figure}[htb]
\centering
  \includegraphics[width=0.8\columnwidth]{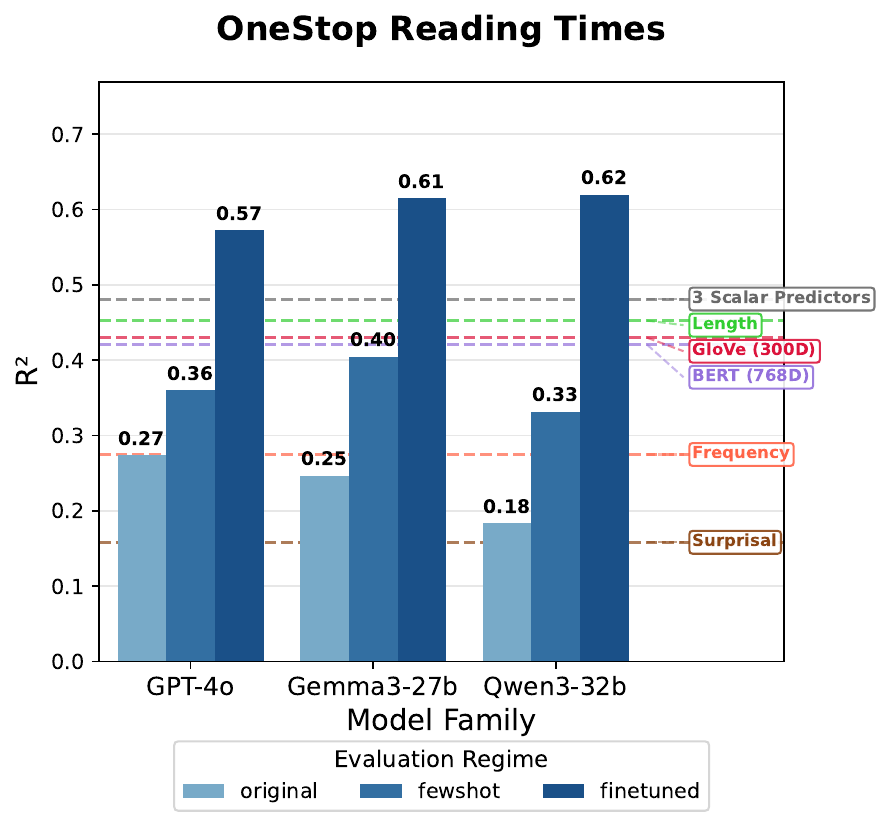}
  \caption {The correlation between model predictions and ground truth norms for eye-tracking reading times (OneStop corpus), across 3 model families and in 3 evaluation regimes. For comparison, the mean correlation with the predictions of interpretable baseline predictors is included.}
  \label{fig:rt-onestop}
\end{figure}

Interestingly, we observe a qualitatively different pattern for eye-tracking reading times from the OneStop Corpus (\ref{fig:rt-onestop}). The zero-shot model attains $R^2$ values of 0.27, the few-shot model attains $R^2$ values of 0.35, while the fine-tuned model attains $R^2$ values in a wide range from  $0.08 \sim 0.57$. For GPT, the fine-tuned model exceeds the predictive power of the baseline predictors, including the combined three scalar predictors of length, frequency, and surprisal, as well as the predictive power of contextual and non-contextual word embeddings. 
For Gemma and Qwen models, predictions from fine-tuning underperform zero-shot and few-shot predictions, indicating a failure of the fine-tuning process to elicit humanlike reading-time predictions. The moderate predictive power of zero-shot and few-shot predictions, however, point to some latent knowledge that correlates with reading times. 

Consistent with known differences between Self-Paced Reading and Eye-Tracking, we find considerable differences in the degree to which the baseline predictors of surprisal, frequency, and length are able to predict reading time measures across the two datasets of Natural Stories and OneStop. 
We note that our baseline results for OneStop are consistent with the work of \cite{opedalRoleContextReading2024}, who found that length and frequency explain a greater deal of variance in reading times, compared to contextual surprisal.


\section{Discussion}

This section discusses the key insights from the experimental results. 

\subsection{LLMs can predict sentence-level norms via fine-tuning}

The task of generating reading times for each word in a sentence is considerably more complex than estimating, for example, the valence of a single word. The same applies to estimating the memorability of a sentence. This is because both of these norms involve considering the ways in which multiple words in a sentence interact with each other. For instance, one of the most memorable sentences in the study of \cite{clarkDistinctiveMeaningMakes2026} is \textit{Does olive oil work for tanning?}; the memorability of this sentence cannot simply be reduced to the sum of word-level features, but involves the compositional meaning of the words in relation to each other.

Despite this challenge, LLMs are capable of producing estimates that correlate with human behavior when fine-tuned on a small set of supervised data. For both memorability and reading times, fine-tuning achieves a correlation with the ground truth that exceeds that of simple, theoretically motivated baseline predictors. 
This is consistent with previous results for word-level norms \cite{conde2025adding}, showing that the same general paradigm can be extended to the sentence level.
This means that LLMs can potentially be used to predict those norms which are more difficult to collect than word-level norms which rely only on subjective judgments. We also note that the improvement is not attributable simply to high performance on words seen during fine-tuning --- for the memorability experiments, the evaluation was conducted on entirely held-out stimuli. For the reading time experiments, all test sentences were held-out, though some individual wordforms were seen in both fine-tuning and testing; model prediction accuracy was high even for words not been seen during fine-tuning. 

Crucially, for most datasets, the $R^2$ value attained by the fine-tuned prompting method exceeds that of both theoretically motivated baseline predictors, and rich semantic embeddings. This demonstrates that the prompting method not only can capture the variance explained by these interpretable predictors, but captures additional variance as well.  

On a theoretical level, this may suggest that the rich distributed representations of words produced by language models encode sufficient information about a diverse range of psycholinguistic behavioral measures, such that fine-tuning on a small dataset enables reliable decoding of these features from the existing model representations (as opposed to learning the relationship from scratch). 

On a practical level, this method offers a way for psycholinguists to develop predictive models that can estimate psycholinguistic norms for held-out data. While this can introduce challenges in interpretability, the approach may be useful in settings such as the development of experimental materials. One possible use case is ensuring matched control stimuli, as has previously been done with the A-Maze variant \cite{boyceMazeMadeEasy2020} of the Maze reading task \cite{forsterMazeTaskMeasuring2009}. 

\subsection{Zero-shot and few-shot performance vary widely across domains}

In our results, we observed that for word and sentence memorability, zero-shot model predictions were essentially completely uncorrelated from human behavior. We speculate that the features which make words and sentence memorable may not be obvious or transparent, consistent with findings from the memorability literature showing that humans themselves are poor judges of memorability in domains such as images \cite{isolaWhatMakesPhotograph2014}. Meanwhile, the features that correlate strongly with reading times are relatively simple and interpretable in comparison --- basic properties of words such as length and frequency, as well as contextual features such as their predictability given preceding sentential context \cite{haleProbabilisticEarleyParser2001, smithEffectWordPredictability2013, wilcoxTestingPredictionsSurprisal2023, wilcoxPredictivePowerNeural2020}, which may be particularly well-aligned with the next-word prediction training objective of LLMs. Thus we speculate that zero-shot prompting of reading times from LLMs is more effective in extracting predictions that align with human behavior, compared to memorability. 

\subsection{Differences between self-paced reading and eye-tracking}

We also observe considerable differences in the performance of models when comparing self-paced reading and eye-tracking. Eye-tracking remains the gold standard for reading time data, because participants are able to read naturalistically and because the paradigm allows both high spatial and temporal resolution. In self-paced reading, by comparison, there are well-established ``spillover'' effects \cite{raynerEyeMovementsReading1998, smithEffectWordPredictability2013} stemming from the rapid pressing of keys and the latency between reading and motor actions. Additionally, the paradigm is somewhat divorced from reading in the wild. Fine-tuning still yields an improvement in predicting self-paced reading times, compared to zero-shot prompting, but the predictive power may be limited by the noise ceiling of the human data. 

\begin{figure}[htb]
\begin{centering}
  \includegraphics[width=0.8\columnwidth]{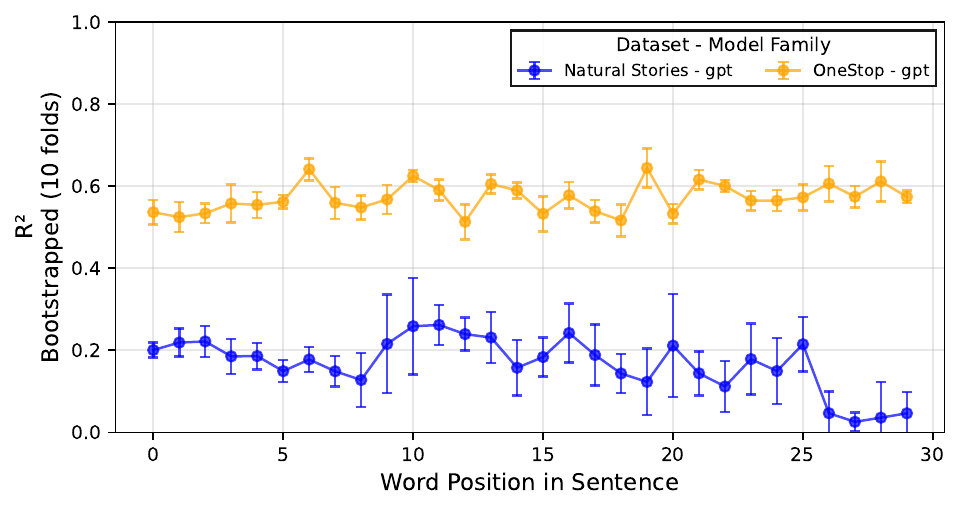} 
  \caption {Correlation of fine-tuned model prediction to ground truth for different word positions in a sentence.}
  \label{fig:corrs-by-word-pos}
  \end{centering}
\end{figure}

\ref{fig:corrs-by-word-pos} shows the $R^2$ values between model-predicted reading times and the ground truth values as a function of word position within sentence, for the NaturalStories and OneStop corpora, using the GPT model. 
For eye-tracking data (OneStop), the $R^2$ between model predictions and the ground truth remains above close to 0.6 even for word positions far into a sentence. 
Meanwhile, for the noisier self-paced reading data, $R^2$ values drop close to 0 for positions beyond approximately 25 words into a sentence. 
We note that the fine-tuning data naturally contains fewer examples of late word positions than early word positions. 
Thus, a single prompt can generate model predictions that correlate with human eye-tracking RTs across a long timescale, but for noisier self-paced reading RTs, model predictions become noisier with word position. 

\section{Conclusion}

In conclusion, this work presents several novel findings. First, this work demonstrates that prompting LLMs for psycholinguistic norms can be extended to sentence-level norms --- the memorability of an entire sentence, or the reading time of a word in context --- not just word-level features, by using a simple and straightforward supervised fine-tuning strategy. We also showed that the reliability of zero-shot LLM predictions for psycholinguistic norms varies considerably by domain, with the predictions for both word and sentence memorability being uncorrelated with empirical norms. Across domains, fine-tuning on a few hundred examples is generally able to align the model predictions with empirical values, even beating strong, theoretically motivated baselines.  

While the applications of this method in psycholinguistics are promising, our results also suggest that practitioners must be wary of trusting the zero-shot predictions of LLMs on psycholinguistic norming tasks \cite{huPromptingNotSubstitute2023}, and careful validation against human data is recommended \cite{conde2025adding}. Future work may target a comprehensive application of this method to other language model families and additional human languages, in order to evaluate the generalizability of these results.

\section*{Limitations}

In this section, we acknowledge several limitations of our study. 

A limitation of the present study is the exclusive focus on English. It remains an open question whether prompting works as well in lower-resource languages or languages with very different linguistic properties, such as agglutinative languages or those with non-alphabetic writing systems. This limitation also reflects the broader WEIRD bias (Western, Educated, Industrialized, Rich, Democratic) highlighted in cognitive science and NLP, where research has tended to focus on a narrow set of languages and populations that are not representative of global diversity.

Finally, we echo existing calls for caution \cite{gaoTakeCautionUsing2025, dentellaSystematicTestingThree2023} regarding the use of LLM outputs as a substitute for, rather than a noisy estimate of, human psycholinguistic norms, especially in zero-shot settings without careful validation against ground-truth values. 


\bibliography{references}

\appendix


\section{Use of Artifacts and Models}
We utilize the \texttt{GPT-4o-mini-2024-07-28} language model via the OpenAI API.

We also utilize existing datasets from \cite{tuckuteIntrinsicallyMemorableWords2025} (MIT License), \cite{clarkDistinctiveMeaningMakes2026} (MIT License), \cite{futrellNaturalStoriesCorpus2021} (CC BY-NC-SA 4.0 license.), and \cite{berzakOneStop360ParticipantEnglish2025} (CC BY 4.0), which are publicly available via GitHub or OSF. We use this data purely for research purposes. 

We acknowledge the use of the ChatGPT and Cursor AI assistants for help with code development. 

\section{Risks}

We do not foresee any risks to this research, as it is focused on the evaluation of existing models using a new approach. 

\end{document}